\documentclass[letterpaper]{article} % DO NOT CHANGE THIS
\usepackage[]{aaai25}  % DO NOT CHANGE THIS
\usepackage{times}  % DO NOT CHANGE THIS
\usepackage{helvet}  % DO NOT CHANGE THIS
\usepackage{courier}  % DO NOT CHANGE THIS
\usepackage[hyphens]{url}  % DO NOT CHANGE THIS
\usepackage{graphicx} % DO NOT CHANGE THIS
\usepackage{natbib}  % DO NOT CHANGE THIS AND DO NOT ADD ANY OPTIONS TO IT
\usepackage{caption} % DO NOT CHANGE THIS AND DO NOT ADD ANY OPTIONS TO IT
\usepackage{algorithm}
\usepackage{algorithmic}
\usepackage{todonotes}
\usepackage{newfloat}
\usepackage{listings}
\DeclareCaptionStyle{ruled}{labelfont=normalfont,labelsep=colon,strut=off} % DO NOT CHANGE THIS
\floatstyle{ruled}
\newfloat{listing}{tb}{lst}{}
\floatname{listing}{Listing}
\usepackage{subfig}
\usepackage{threeparttable}
\usepackage{multirow}
\usepackage{booktabs} 
\usepackage{amssymb}
\usepackage{amsmath}
\usepackage{makecell}
\usepackage{xcolor}
\usepackage{pifont}
\usepackage{colortbl}
\usepackage{adjustbox}
\usepackage{fontawesome}
\definecolor{brandeisblue}{rgb}{0.0, 0.44, 1.0}
\definecolor{mygray}{RGB}{220,220,220}
\definecolor{deepgreen}{rgb}{0,0.6,0}
\usepackage{comment}
\newcommand{\ignore}[1]{}
\usepackage{hyperref}

\title{Identifying Query-Relevant Neurons in Large Language Models \\for Long-Form Texts}
\ignore{
\author{
    Written by AAAI Press Staff\textsuperscript{\rm 1}\thanks{With help from the AAAI Publications Committee.}\\
    AAAI Style Contributions by Pater Patel Schneider,
    Sunil Issar,\\
    J. Scott Penberthy,
    George Ferguson,
    Hans Guesgen,
    Francisco Cruz\equalcontrib,
    Marc Pujol-Gonzalez\equalcontrib
}
\affiliations{
    \textsuperscript{\rm 1}Association for the Advancement of Artificial Intelligence\\
    1101 Pennsylvania Ave, NW Suite 300\\
    Washington, DC 20004 USA\\
    proceedings-questions@aaai.org
}
}

\author{
Lihu Chen, Adam Dejl, Francesca Toni\\
}

\affiliations {
    Imperial College, London, UK\\
    firstname.surname@imperial.ac.uk
}
\usepackage{bibentry}
\begin{document}

\maketitle

\begin{abstract}
Large Language Models (LLMs) possess vast amounts of knowledge within their parameters, prompting research into methods for locating and editing this knowledge. Previous work has largely focused on locating entity-related (often \emph{single-token}) facts in smaller models.
However, several key questions remain unanswered: (1) \textit{How can we effectively locate query-relevant neurons in decoder-only LLMs, such as Llama and Mistral? }
(2) \textit{How can we address the challenge of long-form (or free-form) text generation?}
(3) \textit{Are there localized knowledge regions in LLMs?}
In this study, we introduce Query-Relevant Neuron Cluster Attribution (QRNCA), a novel architecture-agnostic framework capable of identifying query-relevant neurons in LLMs. QRNCA allows for the examination
of long-form answers beyond triplet facts by employing the proxy task of multi-choice question answering. To evaluate the effectiveness of our detected neurons, we
build two multi-choice QA datasets spanning diverse domains and languages.
Empirical evaluations demonstrate that our method outperforms baseline methods significantly. Further, analysis of neuron distributions reveals the presence of visible localized regions, particularly within different domains. Finally, we %demonstrate the
show potential applications of our detected neurons in knowledge editing and neuron-based prediction.
\faGithub~ \url{https://github.com/tigerchen52/qrneuron}
%Extended version \url{https://arxiv.org/abs/2406.10868}
\end{abstract}


\section{Introduction}

Large Language Models (LLMs) contain substantial amounts of knowledge within their neurons (or parameters). 
Recent research has focused on identifying and localizing these knowledge neurons to gain insights into the information processing mechanisms of LLMs. 
\emph{Activation-based methods}~\cite{voita2023neurons} examine activation patterns to elucidate 
the role of neurons in the reasoning process. 
However, these methods often struggle to directly attribute specific outputs to corresponding inputs, thereby limiting their effectiveness in accurately identifying relevant knowledge.
\emph{Gradient-based methods}~\cite{dai2022knowledge} measure the sensitivity of model outputs to internal components in response to specific inputs, which enables the effective identification of neurons relevant to particular queries. However, these methods typically employ fill-in-the-blank tasks, such as ``\texttt{Paris is the capital of \rule{0.08\linewidth}{0.4pt}}'', to localise components representing \emph{triplet facts}.
\emph{Causality-based methods}~\cite{meng2022locating} take a different approach by employing causal mediation analysis to pinpoint \emph{layers} within LLMs that store factual associations.
Another branch of pioneering research attempts to locate functional regions in small-size language models such as BERT~\cite{kenton2019bert} and GPT-small~\cite{radford2019language}, including linguistic regions~\cite{zhang2024unveiling}, factual subnetworks~\cite{ren2022specializing,bayazit2023discovering}, and modular structures~\cite{zhang2023emergent,conmy2023towards}.

While these studies successfully identify knowledge associations stored within LLMs, three significant questions remain underexplored:
(1) How can we effectively locate query-relevant neurons in contemporary decoder-only LLMs, such as Llama~\cite{touvron2023llama} and Mistral~\cite{jiang2023mistral}, given their large model size and different architectures?
(2) How can we address the challenge of long-form text generation, as previous methods have been limited to triplet facts?
(3) Are there localized knowledge regions in LLMs analogous to the localized functional regions observed in human brains~\cite{brett2002problem}?

\begin{table}[t]
\tiny
\centering
  %\fontsize{8}{10}\selectfont 
  \setlength{\tabcolsep}{0.99em}
  \begin{tabular}{l|cccc}
  \toprule
  \bfseries Methods & \bfseries \makecell{Long-Form \\ Texts} & \bfseries \makecell{Neuron-Level \\ Location} & \bfseries \makecell{Decoder \\ Models} & \bfseries \makecell{$\geqslant$ 7B \\ LLMs}\\
  \midrule
  Knowledge Neuron~\shortcite{dai2022knowledge} & \color{red!80}{\ding{55}} & \color{deepgreen}{\ding{51}} & \color{red!80}{\ding{55}} & \color{red!80}{\ding{55}}\\
  ROME~\shortcite{meng2022locating} & \color{red!80}{\ding{55}} & \color{red!80}{\ding{55}} & \color{deepgreen}{\ding{51}} & \color{red!80}{\ding{55}}\\
  Knowledge Subnetwork~\shortcite{bayazit2023discovering} & \color{red!80}{\ding{55}} & \color{deepgreen}{\ding{51}} & \color{deepgreen}{\ding{51}} & \color{red!80}{\ding{55}}\\
  \rowcolor{teal!10} \textbf{
  QRNCA (Ours) } & \color{deepgreen}{\ding{51}} & \color{deepgreen}{\ding{51}} & \color{deepgreen}{\ding{51}} & \color{deepgreen}{\ding{51}} \\
  \bottomrule
  \end{tabular}
  \vspace{-2ex}
  \caption{
  Comparison of general-domain knowledge locating methods. Here, we do not consider task-specific approaches like Language Neuron~\cite{chen2024journey} and Privacy Neuron~\cite{wu2023depn}.\label{tab:intro_comparison}
}
\vspace{-10pt}
\end{table}

\begin{figure*}
    \centering
    \includegraphics[width=0.75\textwidth]{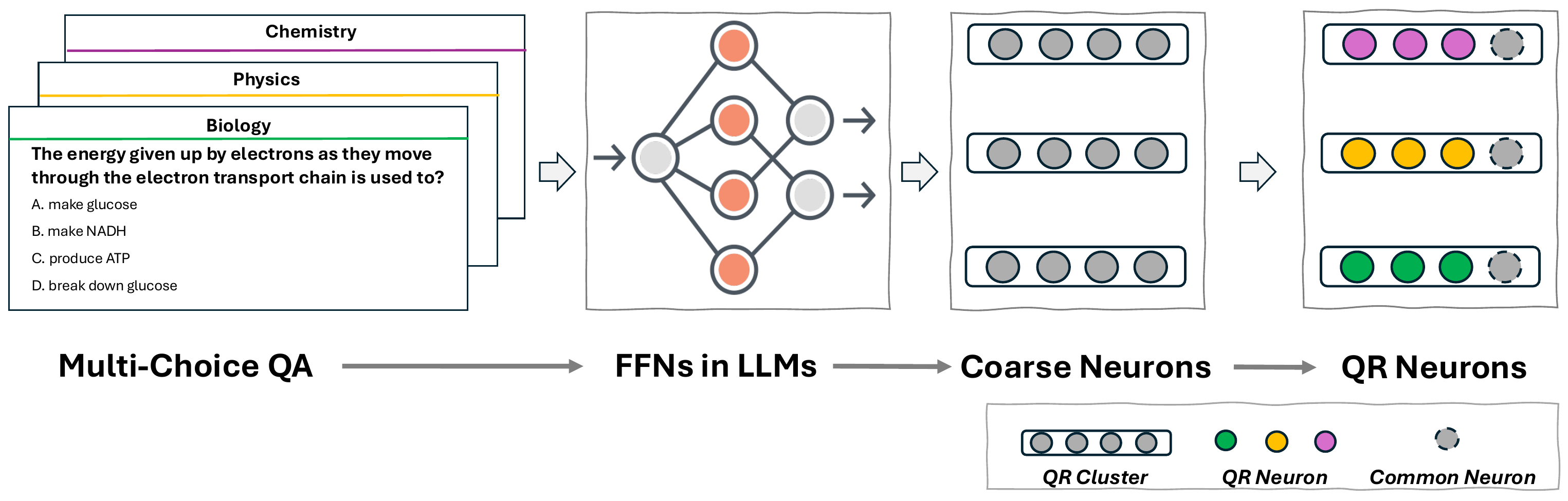}
    \caption{Our overall framework% of our Query-Relevant (QR) Neuron
    , which aims to detect Query-Relevant (QR) neurons with regard to specific queries.}
    \label{fig:framework}
     \vspace{-1ex}
\end{figure*}

To address the first two questions, we introduce a novel framework named \textbf{\emph{Query-Relevant Neuron Cluster Attribution (QRNCA)}} designed to identify query-relevant neurons in LLMs. 
The principal advantages of our framework are its architecture-agnostic nature and its capability of handling long-form text generation effectively, as shown in Table~\ref{tab:intro_comparison}% shows the benefits of our proposed method
.
QRNCA aims to extract Query-Relevant (QR) neurons for each query-answer fact. The process begins by transforming a free-form generation task into a multiple-choice question-answering format. By employing prompt engineering, we constrain LLMs to generate only the option letter rather than the complete answer. This approach allows for the examination of long-form generation beyond single tokens.
Subsequently, we adapt the Knowledge Attribution method~\cite{dai2022knowledge} to compute \emph{Neuron Attribution}, which elucidates the relationship between neurons and the factual knowledge. We then gather clusters for a series of queries and calculate the \emph{Inverse Cluster Attribution}. This step mitigates the influence of neurons that recur across clusters (or queries). The final step involves multiplying the neuron attribution and inverse cluster attribution values to pinpoint correlated neurons.
Additionally, we identify certain \emph{Common Neurons} that are associated with common words, punctuation marks, and option letters. Excluding these common neurons enhances the detection of QR neurons. Empirical evaluations demonstrate that our proposed method outperforms baseline approaches.

To investigate the existence of localized knowledge regions, we construct two multi-choice QA datasets encompassing various \emph{domains} and \emph{languages}. 
Then, we visualize the geographical locations of the detected neurons in Llama. Our findings indicate that distinct localized regions emerge in the middle layers, particularly for domain-specific neurons. This suggests that LLMs tend to complete the formation of domain-specific concepts within these middle layers. 
Conversely, language-specific neurons are more sparsely distributed, indicating that LLMs likely draw on linguistic knowledge at all processing levels. 
Additionally, we observed that common neurons are concentrated in the top layer, predominantly expressing frequently used tokens. 

In summary, our main contribution is four-fold:
(1) \textbf{A scalable method:} we propose QRNCA to detect query-relevant neurons in LLMs; the QRNCA method is architecture-agnostic and can deal with long-form generations. 
(2) \textbf{Two new datasets:} we curate two multi-choice QA datasets that contain different types of knowledge, namely Domain Knowledge and Language knowledge.
(3) \textbf{In-depth studies: } we visualize distributions of detected neurons and we are the first to show that there are visible localized regions in Llama.
(4) \textbf{Potential applications: } we show that QRNCA might be useful for knowledge editing and neuron-based prediction.

\section{Related Work}

\subsection{Locating Knowledge in LLMs}
%Language models 
Large Language Models contain a vast range of knowledge within their parameters, spanning factual~\cite{petroni2019language, zhou2020evaluating, jiang2020x, roberts2020much, pezeshkpour2023measuring}, linguistic~\cite{liu2019linguistic, jawahar2019does, chen2023locality}, and domain-specific information~\cite{sung2021can, frieder2024mathematical}. 
Recent mechanistic studies suggest that knowledge is primarily stored in the FFN (Feed-forward Network) layers of Transformers~\cite{geva2021transformer,geva2022transformer}, which prompts ongoing research efforts aimed at developing methods to identify and locate this knowledge within these layers.
\emph{Activation-based methods}~\cite{voita2023neurons, gurnee2024universal} investigate the activation patterns of neurons to interpret how the network processes information at different stages. However, a key limitation of these methods is their inability to directly attribute the model’s output to specific inputs, which limits their precision in identifying relevant knowledge.
\emph{Gradient-based methods}~\cite{ancona2019gradient,dai2022knowledge}, on the other hand, offer fine-grained attribution by quantifying the sensitivity of model outputs to internal components in response to a given input. This approach effectively identifies neurons relevant to specific queries. Nonetheless, current gradient-based techniques have primarily focused on single-token triplet facts. 
Another approach, \emph{Causality-based methods}, employs causal mediation analysis to discern the particular layers associated with a given factual input~\cite{meng2022locating}. This line of research has evolved into a locate-and-edit paradigm, aimed at refining knowledge within LLMs~\cite{meng2022mass, ju2023klob, zhang2024comprehensive}.
In addition to general knowledge locating approaches, recent studies have focused on identifying neurons responsible for specific tasks, such as linguistic~\cite{chen2024journey,tang2024language,kojima2024multilingual}, privacy-related~\cite{wu2023depn,chen2024learnable} and bias-related neurons~\cite{yang2023crispr}.

In this work, we propose a novel gradient-based attribution method aimed at locating input-output knowledge within LLMs. Unlike previous methodologies, our approach mainly focuses on long-form (or free-form) texts beyond entity facts.

\subsection{Analyzing Knowledge Distribution in LLMs}
Given the human-like reasoning capabilities observed in LLMs across various tasks~\cite{zhao2023survey}, and since our brain contains functional locations associated with distinct cognitive processes~\cite{brett2002problem, bjaalie2002localization,gholipour2007brain}, we ask whether there are similar regions in LLMs.
Previous investigations have explored the behaviors of individual neurons indicating that a neuron can encode multiple concepts~\cite{bolukbasi2021interpretability} while a concept can also be distributed across multiple neurons~\cite{dalvi2019one,durrani2020analyzing, chen2024journey}.
Subsequent endeavors have sought to identify functional regions in LLMs, encompassing linguistic regions~\cite{zhang2024unveiling}, factual subnetworks~\cite{ren2022specializing,bayazit2023discovering}, and modular structures~\cite{zhang2023emergent,conmy2023towards}.
These studies have 
investigated localized behaviors in smaller-scale
language models, such as BERT and GPT-small.
Building upon these foundations, our research embarks on the examination of knowledge locations in larger-size LLMs, specifically those with 7B parameters, spanning multiple knowledge domains.

\section{Background}

\paragraph{Feed-forward Networks in LLMs} Feed-forward networks (FFNs) are widely used by transformer-based language models. \citet{geva2021transformer} reveal that FFNs emulate key-value memories and their outputs are responsible for refining the
final output distribution over the vocabulary. Although traditional two-layer FFNs in BERT~\cite{kenton2019bert} and GPT-2~\cite{radford2019language} have been studied well, the behaviors of FFNs in modern LLMs such as Llama~\cite{touvron2023llama} and Mistral~\cite{jiang2023mistral}, are not well-explored. 
These LLMs adopt Gated Linear Units (GLUs)~\cite{dauphin2017language} in their FFNs, which can be formulated as follows:

\begin{equation}~\label{eq:glus}
	\text{FFN}(\mathbf{X}) = ( \mathbf{X} \mathbf{W}^{U} \odot  \text{SiLU}(\mathbf{X} \mathbf{W}^{G})) ~ \mathbf{W}^{D}
\end{equation}

Here, $\mathbf{X} \in \mathbb{R}^{n \times d}$ is the input sequence, $n$ is the number of tokens and $d$ is the dimension of input vectors; 
$\mathbf{W}^{U} \in \mathbb{R}^{d \times m}$,
$\mathbf{W}^{G} \in \mathbb{R}^{d \times m}$,  $\mathbf{W}^{D} \in \mathbb{R}^{m \times d}$ are parameter matrices, $m$ is the hidden dimension of the FFN and $\odot$ is the Hadamard product; finally
$\text{SiLU}$~\cite{elfwing2018sigmoid} is the activation function.

\paragraph{Knowledge Neurons} \citet{dai2022knowledge} propose a gradient-based \emph{Knowledge Attribution} to identify the knowledge neurons in BERT by using the fill-in-the-blank cloze task. 
Their method evaluates the contribution of each neuron in FFNs to the knowledge predictions. 
Given a query prompt $q$ (``\texttt{Paris is the capital of \rule{0.08\linewidth}{0.4pt}}''), the probability of the
correct answer predicted by %a language model
an LLM can be formulated as:
\begin{equation}
    P_{q}(\hat{w}_{i}^{l}) = p(y^{*}|q, w_{i}^{l} =\hat{w}_{i}^{l})
\end{equation}
where $y^{*}$ is the correct answer (\texttt{France}); $w_{i}^{l}$ denotes the $i$-th intermediate neuron in the $l$-th layer in FFNs;
$\hat{w}_{i}^{l}$ is a constant we assign to $w_{i}^{l}$.

In order to measure the attribution score (or contribution) of a neuron, they gradually change the $w_{i}^{l}$ from 0 to
its original value computed during the forward pass and integrate the gradients~\cite{sundararajan2017axiomatic}:

\begin{equation}
    \text{Attr}(w_{i}^{l}) = \Bar{w}_{i}^{l} \int_{\alpha = 0}^{1} \frac{\partial P_{q}(\alpha \Bar{w}_{i}^{l}) }{\partial w_{i}^{l}}  \mathrm{d}\alpha
\end{equation}
where $\frac{\partial P_{q}(\alpha \Bar{w}_{i}^{l}) }{\partial w_{i}^{l}}$ is the gradient with regard to $w_{i}^{l}$.  $\text{Attr}(\cdot)$ accumulates the output probability change as $\alpha$ gradually varies from 0 to 1. The attribution measures the contribution of the neuron $w_{i}^{l}$ to the correct answer. In practice, the score is estimated by using Riemann Approximation:
\begin{equation}~\label{eq:knowledge_attribution}
    \hat{\text{Attr}}(w_{i}^{l}) = \frac{\Bar{w}_{i}^{l}}{m}  {\textstyle \sum_{k=1}^{m}} \frac{\partial P_{q}(\frac{k}{m} \Bar{w}_{i}^{l}) }{\partial w_{i}^{l}}
\end{equation}
where $m$ is the number of the estimation steps.
Finally, they identify a
coarse set of knowledge neurons whose attribution
scores are greater than a threshold $t$. 
The localized neurons are supposed to be highly associated with a piece of knowledge, i.e., the query-answer facts. 

\section{Locating Query-Relevant (QR) Neurons in Decoder-only LLMs}
While Knowledge Attribution~\cite{dai2022knowledge} effectively identifies neurons linked to factual queries, its applicability is limited to encoder-only architectures, and it mandates the output to be a single-token word. To address these constraints, we propose a new framework named Query-Relevant Neuron Cluster Attribution (QRNCA). The framework is architecture-agnostic and capable of handling long-form generation.

To clarify the main concepts in our framework, we provide the following key %definitions
notions: \emph{\textbf{QR Neuron}} is an individual neuron highly correlated with a specific factual knowledge, capable of influencing the corresponding knowledge expression. 
\emph{\textbf{QR Cluster}} represents a coarse grouping of neurons associated with a specific fact. This cluster may include noisy neurons and require further refinement.
\emph{\textbf{Common Neuron}} is consistently activated by a wide range of inputs, representing general knowledge or concepts.

The overall framework is shown in Figure~\ref{fig:framework}.
Our framework first resorts to the proxy task of \emph{Multi-Choice QA} to deal with long-form texts.
Starting with a given input, the framework employs \emph{Neuron Attribution} to derive a QR Cluster%, which consists of a coarse grouping of neurons
. Each neuron within this cluster is assigned an attribution score that indicates its relevance to the query.
Next, we apply \emph{Inverse Cluster Attribution} to attenuate the influence of neurons that frequently occur across multiple clusters. Finally, we identify \emph{ Common Neurons},  as those lacking discriminative power in determining query relevance and representing common knowledge or concepts%, which we refer to as \emph{Common Neurons}
. Refining the extraction of QR neurons by excluding these common neurons enhances the precision in identifying critical neural correlates.

In the following paragraphs, we introduce the details of these key components in our framework: Multi-Choice QA Transformation, Neuron Attribution, Inverse Cluster Attribution, and Common Neurons.

\subsection{Multi-Choice QA Transformation}

Multi-choice QA is widely used in a variety of real-world educational assessments and standardized tests.
Meanwhile, many known benchmarks such as MMLU~\cite{hendrycks2020measuring} and Big-bench~\cite{srivastava2023beyond} use multi-choice QA to evaluate the breadth and depth of a model’s knowledge. 
Therefore, we adopt the proxy task of multi-choice QA to study the knowledge associations in LLMs.
%Given the biology question \textit{``The energy given up by electrons as they move through the electron transport chain is used to?''}, the correct answer can be the long-form text \textit{``produce ATP''}.  
To deal with free-form answers, we advocate for the transformation of questions and their corresponding answers into a multiple-choice framework, as illustrated in Figure~\ref{fig:framework}. This approach involves the generation of distracted options by randomly sampling answers within the same domain. Following this, the LLM is prompted to produce only the option letter. Subsequently, we investigate the neurons correlated with the input. To mitigate the impact of randomness, we devise multiple prompt templates and systematically shuffle the order of options to prevent the model from learning spurious correlations based on option letters. These prompt templates are detailed in Table~\ref{tab:prompt} in the Supplementary Material in the extended version of this paper\footnote{\url{https://arxiv.org/abs/2406.10868}} (SM in short in the remainder of this paper).

\subsection{Neuron Attribution}

To extend our methodology to Gated Linear Units (GLUs), which comprise two linear transformations followed by a gating mechanism, we adapt the Knowledge Attribution approach (Eq~\ref{eq:knowledge_attribution}). In GLUs, the linear transformations involve computing a linear combination of input features, denoted by $f = \mathbf{X} \mathbf{W}^{U}$. Additionally, the gating mechanism, represented by $g = \text{SiLU}(\mathbf{X} \mathbf{W}^{G})$, determines the extent to which each input component should be forwarded, thereby enabling the model to emphasize important features while suppressing irrelevant ones.
To compute the relevant attribution, we can use either   $\frac{\partial P_{q}}{\partial f}$ or $\frac{\partial P_{q}}{\partial g}$ 
and we choose to use the former since our empirical study shows it can obtain better QR neurons (see details in Figure~\ref{fig:gate_prob_change_domain} in the SM).
Given a query $q$, instantiation using our templates yields a query set $\mathcal{Q} = \{q_1, q_2,\ldots,q_{|\mathcal{Q}|} \} $, and the attribution score of the neuron $n_{i}^{l}$ can be denoted as:
\begin{equation}~\label{eq:knowledge_attribution}
    \text{na}(n_{i}^{l}) = \frac{\textstyle \sum_{j=1}^{|\mathcal{Q}|} \frac{\Bar{f}_{i}^{l}}{m}  {\textstyle \sum_{k=1}^{m}} \frac{\partial P_{q_{j}}(\frac{k}{m} \Bar{f}_{i}^{l}) }{\partial f_{i}^{l}}}{Z}
\end{equation}
Here, the numerator means that we sum up the scores of different instantiated templates together as the initial attribution score. The denominator $Z$ is the normalization factor obtained by summing the initial attribution scores of all neurons. 
Since the number of prompts for each query may vary and the initial attribution scores may be scaled differently, we use normalization to make the attribution scores comparable across queries.

\subsection{Inverse Cluster Attribution}
With the attribution score, we can obtain a list of coarse clusters for each query $\mathcal{C} = \{ c_{1}, c_2,\ldots,c_{|\mathcal{C}|}) \}$, where $c$ is a cluster that consists of neurons whose attribution score is higher than some threshold $t$.
The frequent appearance of some neurons %frequently appear 
across queries of different fields %, it
reveals that they are not critical neurons to the input query. To decrease their impact, we calculate the inverse cluster attribution: 
\begin{equation}
    \text{ica}(n_{i}^{l}) = \log \frac{|\mathcal{C}|}{|\{c: c \in \mathcal{C} ~ \text{and} ~ n_{i}^{l} \in c \}| + 1}
\end{equation}

\subsection{Common Neurons}
We observe that some neurons with a relatively high attribution score are still shared across clusters. Through case studies (as shown in Table~\ref{tab:common_neuron_token}), we demonstrate that they express commonly used concepts such as option letters (``\texttt{A}'' and ``\texttt{B}'') or stop words (``\texttt{and}'' and ``\texttt{the}''). Therefore, we count the frequency of each neuron across clusters. If the frequency is higher than the $u\%$ of total clusters, we assign the given neuron into the common neuron set.

\subsection{Obtaining QR Neurons}
Given a query, the final score of a neuron %can be computed as
is given by:
\begin{equation}
    \text{naica}(n_{i}^{l}) = \text{na}(n_{i}^{l}) \times \text{ica}(n_{i}^{l})
\end{equation}
We select top-$v$ neurons with the highest score from the detected cluster and further remove common neurons to refine the QR neuron set.

\section{Analyzing Detected QR Neurons}

\subsection{Experimental Settings}
\subsubsection{Dataset Construction}

We construct two datasets to locate knowledge neurons that cover two different categories: \emph{subject domains and languages}.

\textbf{\emph{Domain Dataset}} is derived from MMLU~\cite{hendrycks2020measuring}, a multi-choice QA benchmark designed to evaluate models across a wide array of subjects with varying difficulty levels. The subjects encompass traditional disciplines such as mathematics and history, as well as specialized fields like law and ethics. In our study, we select six high school exam subjects from the test set: \textit{Biology, Physics, Chemistry, Mathematics, Computer Science}, and \textit{Geography}.

\textbf{\emph{Language Dataset}} is adapted from Multilingual LAMA~\cite{kassner2021multilingual}, which is a dataset to investigate
knowledge in language models in a multilingual setting covering 53 languages. We select six languages: \textit{Arabic, English, French, Japanese, Russian} and \textit{Chinese}. Each language subset includes queries that cover five different relations: \texttt{birth\_place}, \texttt{employer}, \texttt{instrument}, \texttt{headquarters\_location}, and \texttt{host\_country}.

%To mitigate sensitivity to prompts and option orders, each query is instantiated with multiple distinct templates (as shown in Table~\ref{tab:prompt} in the SM), and the option orders are shuffled each time. 
The statistics of our datasets are shown in Table~\ref{tab:stat_dataset} and examples can be found in Table~\ref{tab:data_examples} in the SM. 

\subsubsection{Metric}~\label{sec:metric}
We modify the values of neurons to observe their impact on knowledge expression. 
For each query, we record the percentage change in the probability of the correct answer, thereby assessing the extent to which the QR neurons influence the predictions of LLMs. 
We compare our approach to other baseline methods and include a control group with an equal size to determine whether the same detected neurons affect the predictions of randomly selected queries from unrelated fields (\emph{Unrelated}).
The Probability Change Ratio (PCR) for a dataset is calculated by $\frac{|\text{Related}|}{|\text{Unrelated}|}$, where  $|\text{Related}|$ and $|\text{Unrelated}|$ mean the average probability change of the related and unrelated samples, respectively. 
We hope that detected neurons can affect the knowledge expressions of the corresponding facts (related) while exerting a low impact on unrelated facts.
A higher value of PCR shows detected neurons can have a higher influence on the query, %which is 
indicating better neurons~\cite{chen2024journey}.

\begin{table}[t] 
	
	\centering
        \scriptsize
	\setlength{\tabcolsep}{1.5mm}{
		\begin{threeparttable} 
			\begin{tabular}{c|c|c|c|c|c|c|c}
                \toprule
				\cellcolor{gray!10}{\textbf{Domain}}&\cellcolor{gray!10}{\textit{Bio}}&\cellcolor{gray!10}{\textit{Phys}}&\cellcolor{gray!10}{\textit{Chem}}&\cellcolor{gray!10}{\textit{Math}}&\cellcolor{gray!10}{\textit{CS}}&\cellcolor{gray!10}{\textit{Geo}}&\cellcolor{gray!10}{\textit{Total}}\cr
				\midrule
                    Num &100&100&100&100&52&100&552\cr
                    \midrule
                    \cellcolor{gray!10}{\textbf{Language}}&\cellcolor{gray!10}{\textit{Ar}}&\cellcolor{gray!10}{\textit{En}}&\cellcolor{gray!10}{\textit{Fr}}&\cellcolor{gray!10}{\textit{Ja}}&\cellcolor{gray!10}{\textit{Ru}}&\cellcolor{gray!10}{\textit{Zh}}&\cellcolor{gray!10}{\textit{Total}}\cr
				\midrule
                    Num &100&100&100&100&100&100&600\cr
				\bottomrule
			\end{tabular}
			\caption{Statistics of our constructed datasets. } 	\label{tab:stat_dataset}
		\end{threeparttable}
	}
	%\end{minipage}%
\end{table}

\begin{figure*}[t]%
	\centering
	\subfloat[\centering Overlap Rate]{{\includegraphics[width=0.55\textwidth]{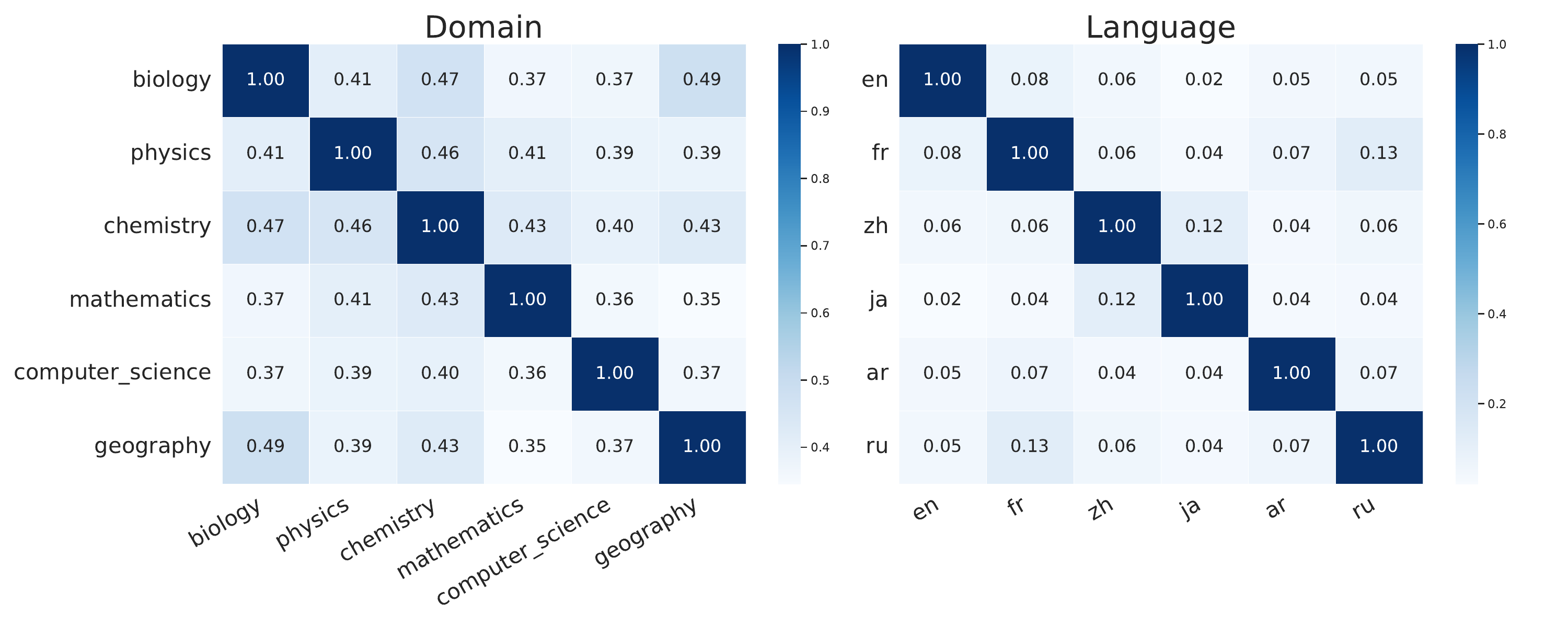} }~\label{fig:overlap_rate}}%
	%\qquad
	\subfloat[\centering Layer Distribution]{{\includegraphics[width=0.3\textwidth]{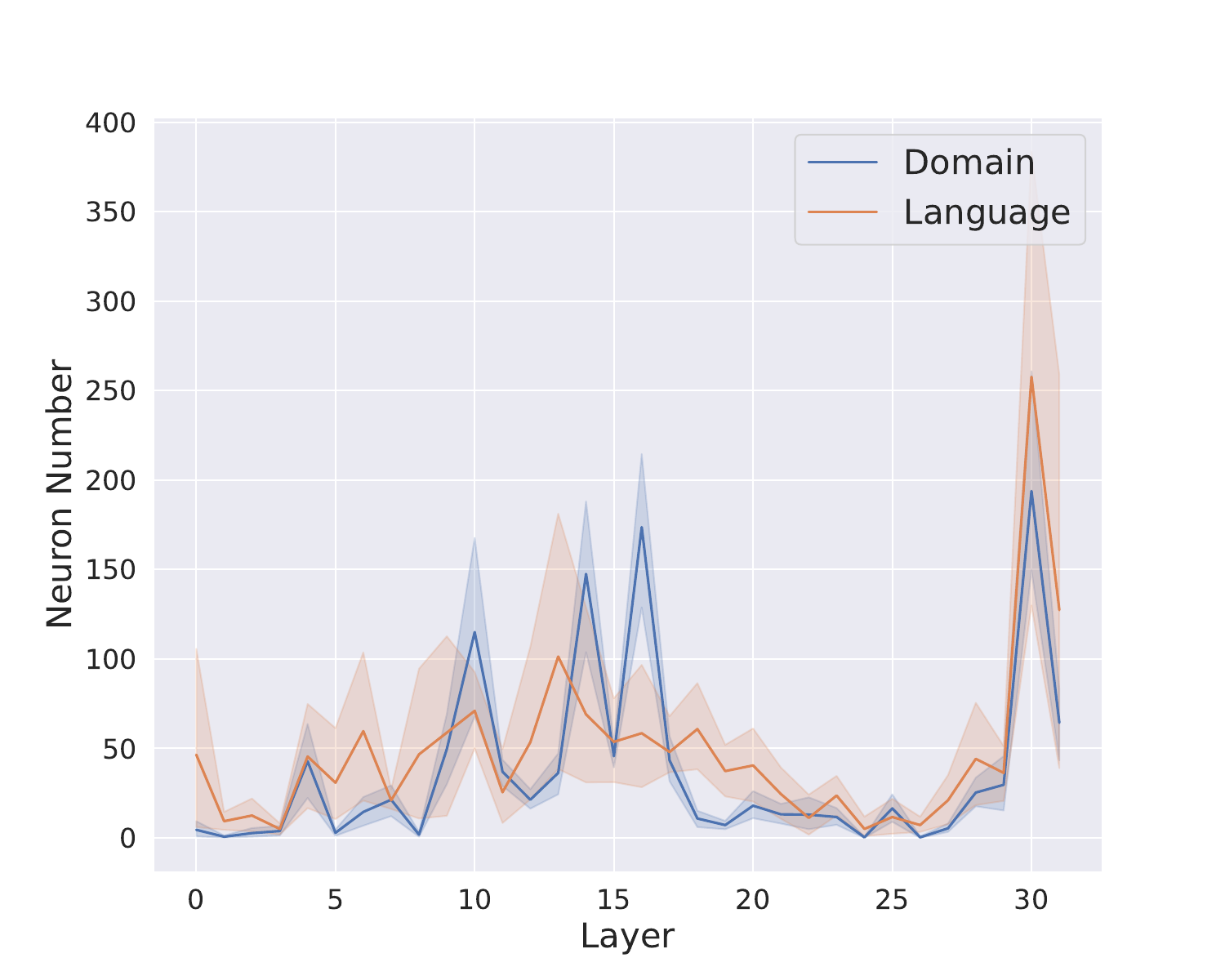} }~\label{fig:layer_distribution}}%
	\caption{Overlap rates and layer distributions of found QR neurons. }
	\label{fig:overlap_distribution}%
\end{figure*}

\subsubsection{Baselines}
We compare %our 
QRNCA to other neuron-level baselines\footnote{We do not compare to ROME~\cite{meng2022locating} since it locates layers instead of neurons. Also, we do not compare to task-specific methods.}: \textbf{Random Neuron} are randomly selected from FFNs% and we make 
, making sure they have the same number of neurons as QRNCA; \textbf{Activation} selects neurons with high activated values. \textbf{Kowledge Neuron}$^*$ is adapted from knowledge attribution~\cite{dai2022knowledge} by using the multi-choice QA task;
\textbf{QRNCA \textit{wo/} ICA} only uses neuron attribution (Eq~\ref{eq:knowledge_attribution}) to obtain relevant neurons, which dose not involve the computation of Inverse Cluster Attribution;
\textbf{QRNCA \textit{w/} Common Neuron} is a variant without removing common neurons.

\begin{table}[t] 
	\centering
        \scriptsize
	\setlength{\tabcolsep}{1.2mm}{
		\begin{threeparttable} 
			\begin{tabular}{c|cc|cc}
                \toprule
                    &\multicolumn{2}{c|}{\textbf{\underline{\texttt{Domain}}}}&\multicolumn{2}{c}{\textbf{\underline{\texttt{Language}}}}\cr
                    \midrule
				Method&\textcolor{brandeisblue}{$\Uparrow$} Boost& \textcolor{brandeisblue}{$\Uparrow$} Suppress&\textcolor{brandeisblue}{$\Uparrow$} Boost& \textcolor{brandeisblue}{$\Uparrow$} Suppress\cr
				\midrule
                    Random Neuron &1.0&0.55&2.0&1.0\cr
                    Activation &1.0&1.0&1.1&1.1\cr
                    Knowledge Neuron$^*$ &1.0&1.0&6.7&1.8\cr
                    QRNCA \textit{wo/} ICA &2.5&1.1&6.5&2.2\cr
                    QRNCA \textit{w/} Common Neuron&2.8&1.8&10.4&8.5\cr
\rowcolor{teal!5}QRNCA&\textbf{4.4}&\textbf{5.6}&\textbf{41.2}&\textbf{36.0}\cr
				\bottomrule
			\end{tabular}
			\caption{Comparisons of different knowledge locating methods for Llama-2-7B. The metric here is the Probability Change Ratio (PCR) described in Section~\ref{sec:metric}. Details are shown in Table~\ref{tab:detail_prob_impact} in the %appendix
            SM.} \label{tab:overall_impact}
		\end{threeparttable}
	}
	%\end{minipage}%
	\vspace{-10pt}
\end{table}

\subsubsection{Implementations}
We mainly study the knowledge neurons in Llama-2-7B~\cite{touvron2023llama} and we 
use the instruction-tuned version so that the model is more responsive to our prompts. 
Llama-2-7B consists of 32 layers with the FFN hidden dimension of 11008.
Besides, we also conduct experiments for Mistral-7B~\cite{jiang2023mistral} to validate whether our method can obtain consistent findings over different models. Note that our framework can be easily extended to larger-size LLMs.  

As for the %used 
hyper-parameters, the number of estimation steps was set to $m \!=\! 16$ and the attribution threshold $t$ to 0.2 times the
maximum attribution score. The template number was $|\mathcal{Q}| \!=\! 3$, the frequency $u$ for obtaining common neurons was 30\%, and the top-$v$ for select coarse neurons was 20. We ran all experiments on three NVIDIA-V100. It took 120 seconds on average to locate neurons for a query with three prompt templates.
For each domain and language, the average number of detected QR neurons is between 12 and 17 (%as shown in
see Table~\ref{tab:stat_key_neurons} in the SM).
Hyper-parameters are selected based on a hold-out set of biology queries with 50 samples.

\subsection{Statistics of Detected QR Neurons}
We are curious about the distribution of different knowledge storage in neurons: \textit{Do different categories of knowledge share neurons?}
To this end, we study the overlap rate. First, we aggregate detected neurons of all queries in a domain or language. Next, the rate is obtained by counting the number of shared neurons between different domains or languages. 
%The overlap rate can show the p
Figure~\ref{fig:overlap_rate} illustrates the overlap rates among different domains and languages.  
We observe that interdisciplinary or interconnected languages share a higher overlap rate such as (\textit{geography, biology}) and (\textit{Chinese, Japanese}), which is in line with our intuition.
A surprising finding is that domains have higher overlap rates than languages, which indicates that LLMs tend to allow the storage of multiple domain-specific concepts in a single neuron (polysemantic). Although language-specific neurons are not monosemantic~\cite{chen2024journey}, they prefer to encode one specific language concepts, which is also consistent with recent findings~\cite{tang2024language}.  

Regarding layer distribution, the QR neurons are predominantly located in the middle layers (15-18) and the top layers (around 30), as depicted in Figure~\ref{fig:layer_distribution}. This finding indicates knowledge concepts are mainly stored in the middle and top layers, and we may only modify these neurons for efficient knowledge updating~\cite{ding2023parameter}.

\subsection{QR Neurons Can Impact the Knowledge Expression}~\label{sec:knowledge_expression}
To validate the impact of our identified QR neurons, we replicate the experiments by ~\citet{dai2022knowledge}, updating the values of QR neurons using two methods: given a query and the value of $\bar{f}_{i}^{l}$, we either (1) boost the neurons by doubling
the value $f_{i}^{l} = 2 \times \bar{f}_{i}^{l}$; or (2) suppress the neuron by making $f_{i}^{l} = 0$.
After one operation, we record the PCR on a specific dataset to show the quality of these neurons. 

Table~\ref{tab:overall_impact} presents the overall performance of various methods. Our QRNCA method consistently outperforms other baselines, evidenced by its higher PCR.
This indicates that our identified QR neurons significantly affect the probability of correct answers while exerting a relatively low impact on unrelated queries. 
For instance, our method achieves a boosting ratio of 41.2 on the language dataset, the highest among the baselines. Additionally, both our proposed ICA and the removal of common neurons provide further benefits in locating neurons, as evidenced by the worse performance of the two QRNCA variants.

\begin{figure}[t]%
	\centering
      \subfloat[\centering Languages]{{\includegraphics[width=0.35\textwidth]{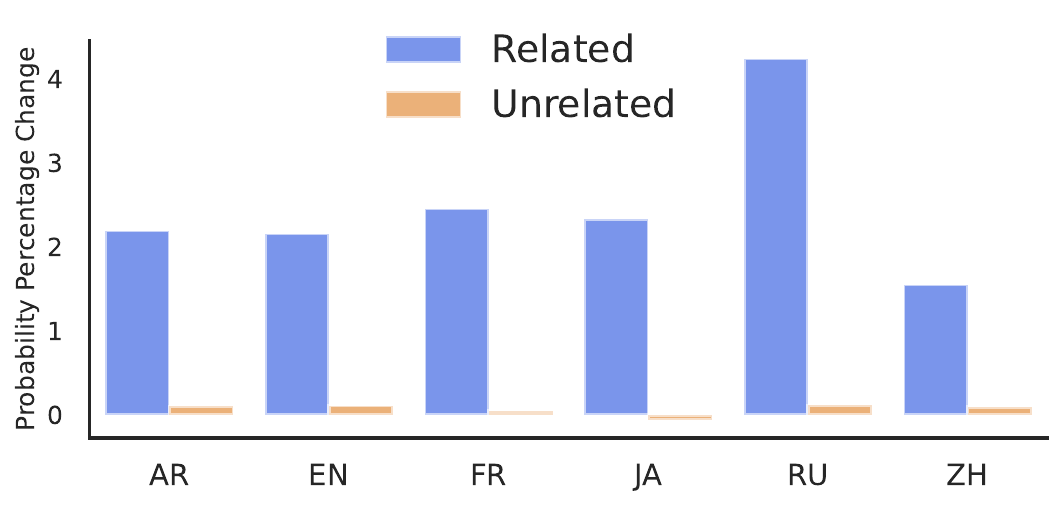} }}%
      \qquad
        \subfloat[\centering  Domains]{{\includegraphics[width=0.35\textwidth]{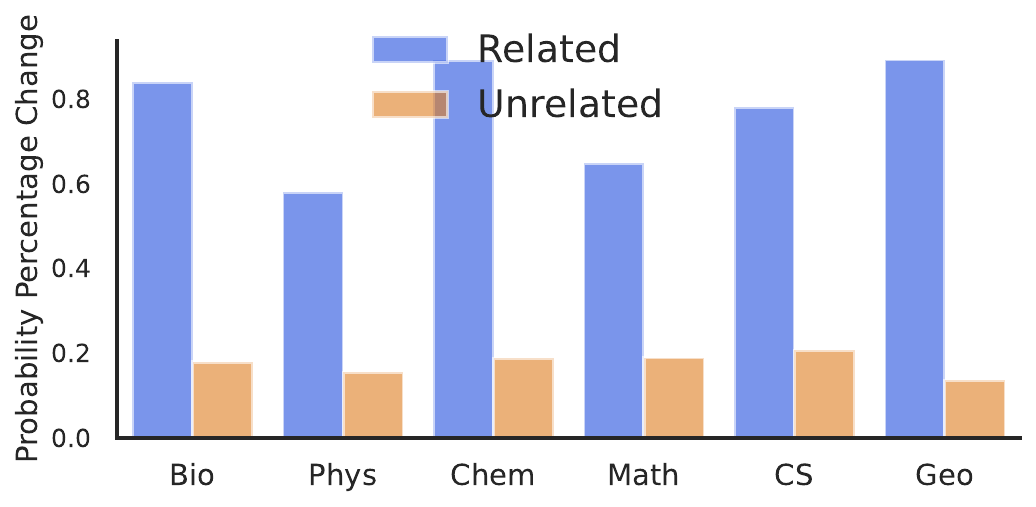} }}%
	\qquad
	\caption{The correct probability percentage change by boosting QR neurons. The LLM here is Llama-2-7B~\cite{touvron2023llama}. The suppression results are shown in Figure~\ref{fig:llama_prob_change_supress} in the SM.} 
	\label{fig:llama_prob_change}%
 \vspace{-10pt}
\end{figure}

Furthermore, Figure~\ref{fig:llama_prob_change} illustrates the percentage change in probability for each domain and language after boosting neuron values. Again, we can clearly observe the effectiveness of our detected QR neurons.
Additionally, we performed %supplementary 
experiments on Mistral-7B. The results, presented in Figure~\ref{fig:mistral_prob_change_domain} in
%\cite{arxiv}
the SM, consistently support our conclusions.

\begin{figure*}
    \centering
    \includegraphics[width=0.8\textwidth]{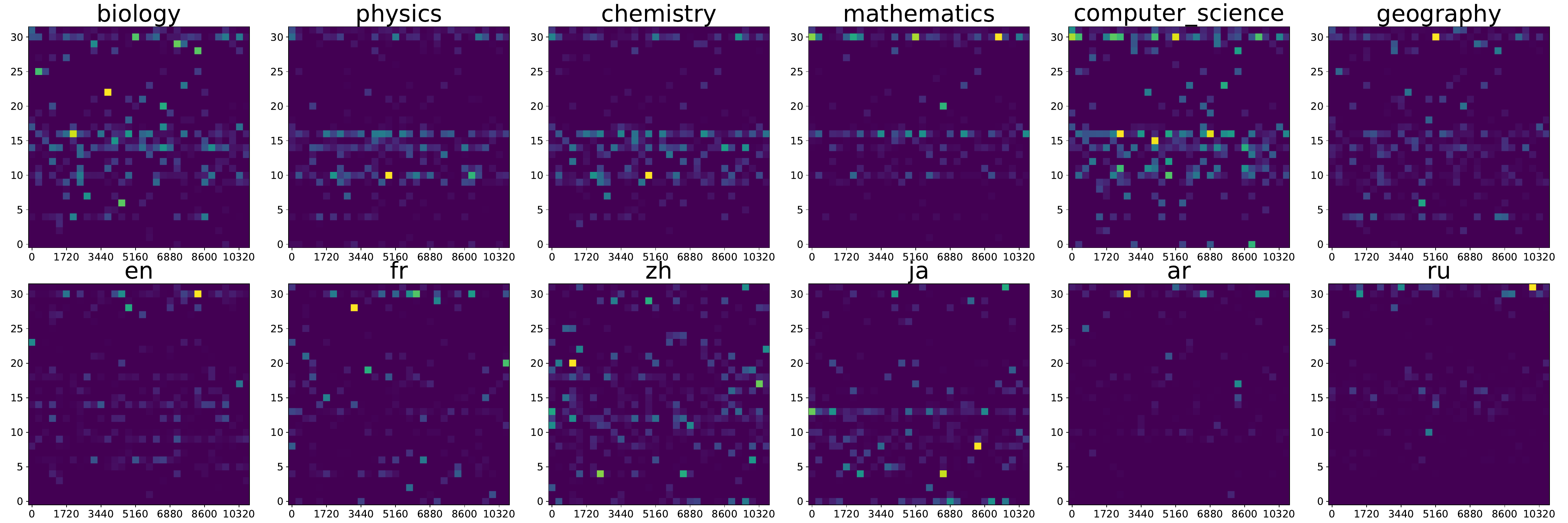}
    \caption{Geographical heatmap of detected QR neurons for different domains and languages. The value is calculated by our $\text{naica}(n_{i}^{l})$. Brighter colors indicate higher $\text{naica}$ values.  The LLM here is Llama-2-7B (11008 $\times$ 32)~\cite{touvron2023llama}}
    \label{fig:location_heat_map}
    \vspace{-10pt}
\end{figure*}

\subsection{Are There Localized Regions in LLMs?}~\label{sec:geographical_location}
Given our ability to identify QR neurons for each query, it is intriguing to explore whether LLMs exhibit localized regions for each domain or language, analogous to the functional localizations in the human brain~\cite{brett2002problem}. 
To investigate this, we visualize domain- or language-specific neurons on a 2D geographical heatmap. The width of the heatmap corresponds to the dimension of FFNs in Llama-2-7B (11008), and the length represents the layer depth (32). We accumulate the value of $\text{naica}(n_{i}^{l})$ to populate the heatmap.
Figure~\ref{fig:location_heat_map} displays the geographical locations of QR neurons in Llama-2-7B across various academic domains and languages. The distribution of QR neurons appears sparse but with distinct regions, particularly for different domains. Notably, certain regions are visible in the middle layers (10-15), suggesting specific neuron patterns. In contrast, language neurons are more sparsely distributed with smaller regions, and languages like Arabic and Russian exhibit less localized properties. 

Based on prior studies, LLMs process and represent information in a hierarchical manner~\cite{geva2022transformer, wendler2024llamas, tang2024language}. The early layers are primarily responsible for extracting low-level features %from the input text
, while the middle layers begin to integrate this information, forming more complex semantic representations. The late layers are typically dedicated to generating the final output.
Therefore, we suppose that domain-specific knowledge representation is built in the middle layer and the top layers are then mainly responsible for next-token prediction, which may explain the visible regions for different subject domains.
Regarding language-specific neurons, their role in accessing linguistic knowledge across different layers likely accounts for their more sparse and distributed locations. This distribution reflects the necessity of engaging with language-specific neurons at multiple stages of information processing.

\subsection{The Function of Common Neurons}
To gain insights into the function of common neurons, we project the matrix $\mathbf{W}^{D}$ in Equation~\ref{eq:glus} to the vocabulary space and select the top-k tokens with the highest probability. 
Table~\ref{tab:common_neuron_token}  lists the predicted tokens, which include common words, punctuation marks, and option letters. These findings reinforce the notion that common neurons are not critical for specific queries.
We also visualize their locations within Llama-2-7B and we observe that they tend to appear at the top layer (as shown in Figure~\ref{fig:meta_location_heat_map} in the SM).  

We also analyzed the token predicted by QR neurons, but we found that middle-layer neurons do not have a clear semantic meaning and human-readable concepts mostly appear in the top layer~\cite{wendler2024llamas}. In Section~\ref{sec:neuron-embedding-analysis} in the SM we conduct  semantic meaning analyses of neurons.

\begin{table}[t] 
	\centering
        \tiny
	\setlength{\tabcolsep}{1.8mm}{
		\begin{threeparttable} 
			\begin{tabular}{c|cc|cc}  
				\toprule  
				Neuron&Top-k tokens\cr
				\midrule
                    $n_{2725}^{31}$ &\texttt{\_in, \_and, \_to, \_for, \_today, \_at, \_as }\cr
                    $n_{10676}^{31}$ &\texttt{\_July, \_June, \_March, \_April, \_November}\cr
                    $n_{10075}^{30}$ &\texttt{., \_, (, :, ), [, -}\cr
                    $n_{5202}^{31}$ &\texttt{\_respectively, \_while, \_and}\cr
                    $n_{5778}^{31}$ &\texttt{\_C, C, \_c, c, '\_ced'}\cr
                    $n_{7670}^{31}$ &\texttt{\_B, B, \_Bill, \_Bh, '\_Bureau'}\cr
				\bottomrule
			\end{tabular}
			\caption{Tokens predicted by the 
            common neurons. } 	\label{tab:common_neuron_token}
		\end{threeparttable}
	}
	%\end{minipage}%
 \vspace{-10pt}
\end{table}

\begin{table}[t] 
	\centering
        \tiny
	\setlength{\tabcolsep}{2.3mm}{
		\begin{threeparttable} 
			\begin{tabular}{c|cc|cc}
                \toprule
                    &\multicolumn{2}{c|}{\textbf{\underline{\texttt{Domain}}}}&\multicolumn{2}{c}{\textbf{\underline{\texttt{Language}}}}\cr
                    \midrule
				Method&Boost& Suppress& Boost&Suppress\cr
                & $\Delta$ (\%) &  $\Delta$ (\%)& $\Delta$ (\%)&   $\Delta$ (\%)\cr
				\midrule
                    Random Neuron &0.0&0.3&0.2&0.3\cr
                    Activation &0.0&0.1&0.0&0.3\cr
                    Knowledge Neuron$^*$ &1.4&3.8&14.3&16.0\cr
                    \rowcolor{teal!5}QRNCA&\textbf{ 12.6}&\textbf{18.2}&\textbf{16.6}&\textbf{24.8}\cr
				\bottomrule
			\end{tabular}
			\caption{Successful rates of knowledge editing. $\Delta$ measures how well we can flip the predictions (\textit{correct} $\rightarrow$ \textit{incorrect} or vice versa).} \label{tab:knowledge_editing}
		\end{threeparttable}
	}
	%\end{minipage}%
	\vspace{-10pt}
\end{table}

\begin{table}[t] 
	\centering
        \tiny
	\setlength{\tabcolsep}{2.2mm}{
		\begin{threeparttable} 
			\begin{tabular}{c|c|c|c}  
				\toprule 
                    Method & \textbf{\underline{\texttt{Biology}}}  &\textbf{\underline{\texttt{Chemistry }}} &\textbf{\underline{\texttt{Geography }}}\cr
				\midrule
                    Random guess &0.25&0.25&0.25\cr
                    Prompt-based model pred. & 0.96 & 0.71 & 0.89 \cr
                    Neuron-based pred. & 0.96 & 0.67 & 0.89 \cr
				\bottomrule
			\end{tabular}
			\caption{Accuracy of neuron-based prediction on selected domains in comparison with the standard prompt-based model prediction. } 	\label{tab:neuron_based_prediction}
		\end{threeparttable}
	}
	%\end{minipage}%
 \vspace{-10pt}
\end{table}

\section{Potential Applications}
We provide two usage examples to showcase the potential applications of our detected QR neurons: \emph{Knowledge Editing} and \emph{Neuron-Based Prediction}.

\subsection{Knowledge Editing}
Apart from using the metric of PCR in Section~\ref{sec:knowledge_expression}, we are also interested in whether the detected QR neurons can be used for knowledge editing.
For this goal, we adjust the values of QR neurons by either boosting or suppressing them to determine if we can change the prediction of a query from incorrect to correct or vice versa. 
Table~\ref{tab:knowledge_editing} presents the success rates of knowledge editing on our constructed language datasets. Our observations indicate that QRNCA achieves higher success rates than other baselines.

\subsection{Neuron-Based Prediction}
The intuition behind neuron-based prediction is that for a domain-specific question, if the corresponding localized regions are properly activated, the LLM is more likely to generate truthful answers. Otherwise, the LLM may produce hallucinated answers.
To this end, we test whether the correct answers to domain-specific questions can be predicted solely based on the activity of the associated neurons. 
Since we harvest QR neurons for queries in different subject domains, we can group all neurons for a domain to obtain a set of \emph{domain-specific neurons}.
We experiment on a specifically constructed MMLU~\cite{hendrycks2020measuring} validation set with a different set of questions than those used to determine the QR neurons (see Section~\ref{sec:neuron-based-prediction-details} in the SM for details on our experimental strategy). 
The results are summarised in Table \ref{tab:neuron_based_prediction}. We observe that the accuracy of the neuron-based predictions is very close to the accuracy of the prompt-based method of using the entire model (the used templates are shown in Table~\ref{tab:prompt} in the SM). This suggests that the activity of identified neurons can reflect the model's reasoning process to some extent. Investigating how this finding could be leveraged in applications like fact-checking and hallucination detection presents a promising line of future work.

\section{Conclusion}
In this study, we introduce a novel framework, QRNCA, for identifying neurons in LLMs for long-form answers, extending beyond triplet facts. 
To validate our approach, we curate two datasets encompassing diverse domains and languages. Our experimental results show that our method outperforms existing baselines in identifying associated neurons. Additionally, this study pioneers the exploration of localized knowledge regions in LLMs and demonstrates Llama contains knowledge-specific regions in the middle layers while language-specific neurons tend to be distributed across different layers.
Further, we prototype two potential usages of identified neurons in applications such as knowledge editing and neuron-based prediction. 
We hope that our findings are beneficial for further research 
in understanding the knowledge mechanisms %of 
underlying LLMs.

\section*{Acknowledgments}
This research was partially supported by the UKRI INDICATE project (Grant No. EP/Y017749/1), by the ERC under the EU’s Horizon 2020 research and innovation program (grant agreement No. 101020934, ADIX), and by J.P. Morgan and the Royal Academy of Engineering under the Research Chairs and Senior Research Fellowships scheme.

\bibliography{aaai25}

\clearpage
\newpage
\appendix
\setcounter{table}{0}   
\setcounter{figure}{0}
\renewcommand{\thetable}{A\arabic{table}}
\renewcommand{\thefigure}{A\arabic{figure}}
\setcounter{equation}{0}
\setcounter{subsection}{0}
\renewcommand{\theequation}{A.\arabic{equation}}

\section{Semantic Analysis of Neurons}
\label{sec:neuron-embedding-analysis}
According to the previous study, Logit Lens\footnote{\url{https://www.lesswrong.com/posts/AcKRB8wDpdaN6v6ru/interpreting-gpt-the-logit-lens}}, the vocabulary probabilistic predictions are a linear function of the activations in Transformer's final layer but we can obtain reasonable distributions if we apply the same function to the activations of intermediate layers, i.e., an interpretable next-token distribution can be obtained by intermediate states.
This finding suggests that intermediate states are capable of representing specific semantic meanings. In Section~\ref{sec:geographical_location}
, we focused on examining the geographical distribution of domain-specific neurons but did not consider their semantic positions. Consequently, a natural question arises:  \emph{what are the properties of the memory cells associated with the QR neurons for the different domains and if they are clustered in the corresponding semantic space}.

\begin{figure}[tb]
    \centering
    \includegraphics[width=0.48\textwidth]{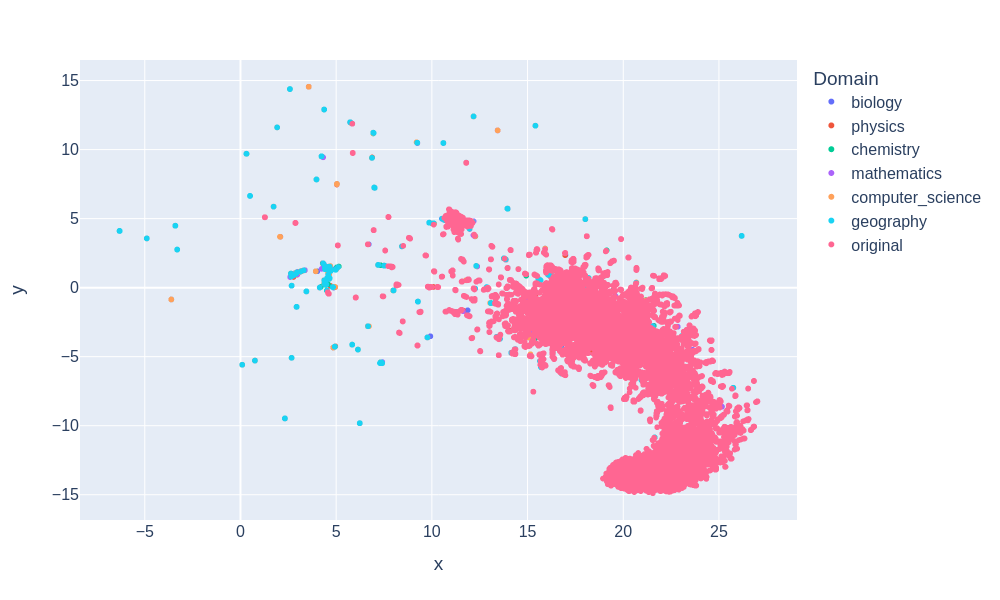}
    \caption{UMAP visualisation of $\mathbf{W}^{D}$ vectors associated with the QR neurons and the token unembeddings}
    \label{fig:latent-value-analysis}
\end{figure}

To this end, we study the hidden activations of $\mathbf{W}^{D}$ (Eq~\ref{eq:glus}), since transformer feed-forward layers can be viewed as key-value memory units.
As a first step in our analysis, we visualize the $\mathbf{W}^{D}$ vectors associated with the QR neurons from the different domains using UMAP \citep{mcinnes2018umap} for dimensionality reduction (with cosine similarity used as the distance metric). For comparison, we additionally include the vectors from the unembedding matrix. The results are shown in Figure \ref{fig:latent-value-analysis}. As can be seen from the figure, the distribution of the vectors associated with QR neurons appears to be significantly different from that of vector unembeddings. Thus, it appears that the contents of the internal memory cells used by Llama 2 are not directly aligned with the final output space. This indicates that Llama 2 tends to form an abstract representation, \emph{usually human-unreadable}, in intermediate layers~\cite{wendler2024llamas}. 

Since the 2D visualization produced by UMAP might not accurately reflect the true properties of the data manifold, we additionally examined the highest-likelihood tokens predicted by QR neurons. Domain-specific neurons are mainly centralized in middle layers, and we found the predicted tokens less human-interpretable, including tokens like \texttt{textt}, \texttt{archivi}, \texttt{\_Kontrola}, \texttt{\_totalité} or \texttt{\_Einzeln}.  Apart from the above tokens, there are certain neurons scattered in top layers still representing option letters, which need further refinement. In summary, since the detected neurons centralize in middle layers, it is hard to interpret their predicted tokens. We may need to explore a better semantic space to study their localized regions.

\begin{table}[b] 
	
	\centering
        \small
	\setlength{\tabcolsep}{1.5mm}{
		\begin{threeparttable} 
			\begin{tabular}{c|c|c|c|c|c|c|c}
                \toprule
				\cellcolor{gray!10}{\textbf{Domain}}&\cellcolor{gray!10}{\textit{Bio}}&\cellcolor{gray!10}{\textit{Phys}}&\cellcolor{gray!10}{\textit{Chem}}&\cellcolor{gray!10}{\textit{Math}}&\cellcolor{gray!10}{\textit{CS}}&\cellcolor{gray!10}{\textit{Geo}}&\cellcolor{gray!10}{\textit{Total}}\cr
				\midrule
                    Num &13.1&13.3&12.8&11.1&14.3&12.7&12.9\cr
                    \midrule
                    \cellcolor{gray!10}{\textbf{Language}}&\cellcolor{gray!10}{\textit{Ar}}&\cellcolor{gray!10}{\textit{En}}&\cellcolor{gray!10}{\textit{Fr}}&\cellcolor{gray!10}{\textit{Ja}}&\cellcolor{gray!10}{\textit{Ru}}&\cellcolor{gray!10}{\textit{Zh}}&\cellcolor{gray!10}{\textit{Total}}\cr
				\midrule
                    Num &12.4&14.4&12.7&16.6&15.8&15.0&15.2\cr
				\bottomrule
			\end{tabular}
			\caption{Average number of detected QR neurons per query.} 	\label{tab:stat_key_neurons}
		\end{threeparttable}
	}
	%\end{minipage}%
\end{table}

\begin{figure}
    \centering
    \includegraphics[width=0.45\textwidth]{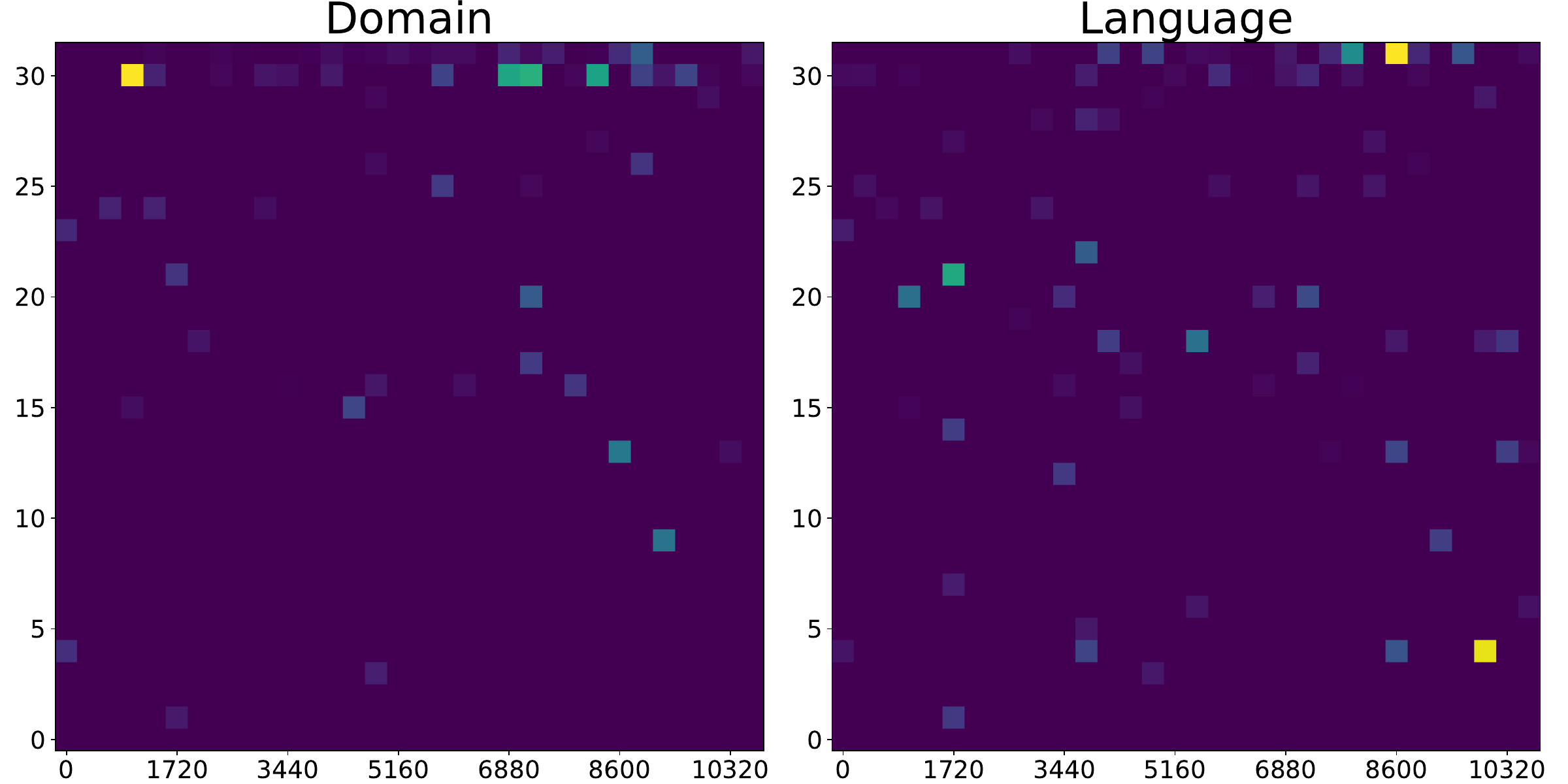}
    \caption{The distribution of common neurons.}
    \label{fig:meta_location_heat_map}
\end{figure}

\section{Neuron-Based Prediction Details}
\label{sec:neuron-based-prediction-details}
In the neuron-based prediction case study, we experiment on the MMLU~\cite{hendrycks2020measuring} validation set $\mathcal{D}_{val}$ to ensure there is no overlap between the dataset used to mine the QR neurons and the test set $\mathcal{D}_{test}$.  As a further post-processing step, we randomly select three options from other domains to replace the incorrect options in each query. Additionally, we manually remove questions that become invalid due to this post-processing, including queries such as \textit{``Which of the following is LEAST valid?''}  and \textit{``All of the following statements are true EXCEPT''}. These operations result in $\sim$20 test samples per domain.
To obtain \emph{domain-specific neurons}, detected QR neurons for each query in a particular domain are grouped together, and we hope the activation of these neurons can be an indicator for predicting the correct answer.
To perform the neuron-based prediction, we compute the gradient of the probability of each option token with respect to the QR neurons for the domain of the considered query, and select the option with the highest total gradient. For comparison, we include a normal prompt-based prediction, which employs designed prompts to query LLM without accessing the internal states (used prompts are shown in Table~\ref{tab:prompt}).

\begin{figure*}[t]%
	\centering
	% \subfloat[\centering Boosting key neurons of domains]{{\includegraphics[width=0.5\textwidth]{images/domian_enhance.pdf} }}%
	%\qquad
	\subfloat[\centering  Domains]{{\includegraphics[width=0.48\textwidth]{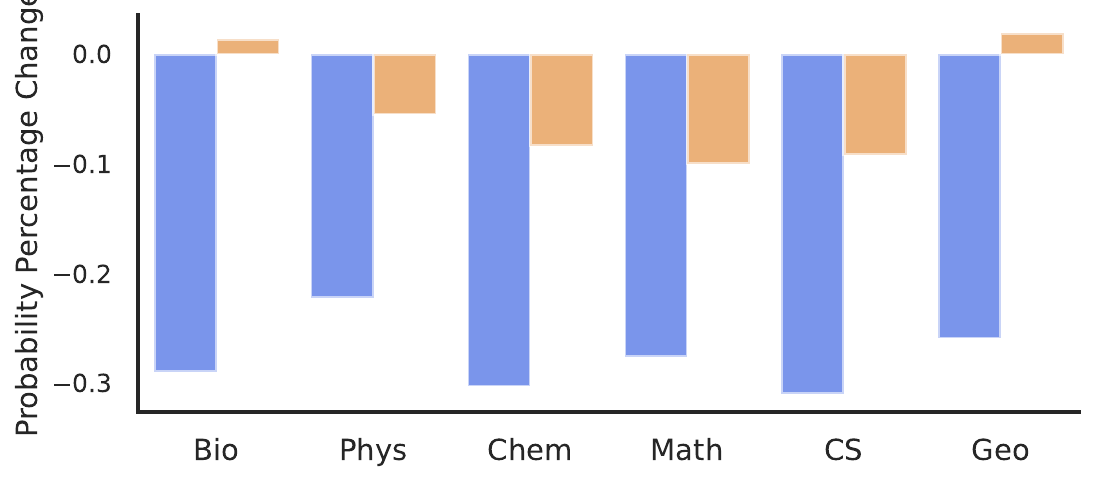} }}%
        \subfloat[\centering Languages]{{\includegraphics[width=0.5\textwidth]{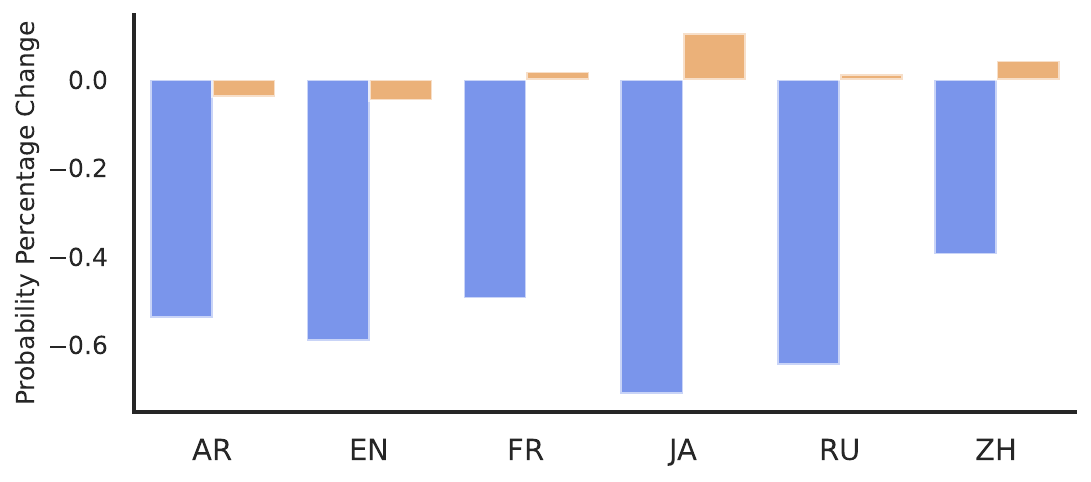} }}%
 
	\caption{The correct probability percentage change after \textbf{\textcolor{red}{suppressing}}. The LLM here is Llama-2-7B~\cite{touvron2023llama}}
	\label{fig:llama_prob_change_supress}%
 \vspace{-10pt}
\end{figure*}

\begin{table*}[t] 
	
	\centering
        \tiny
	\setlength{\tabcolsep}{1.3mm}{
		\begin{threeparttable} 
			\begin{tabular}{c|ccc|ccc|ccc|ccc}
                \toprule
                    &\multicolumn{6}{c|}{\textbf{\underline{\texttt{Domain}}}}&\multicolumn{6}{c}{\textbf{\underline{\texttt{Language}}}}\cr
                    \midrule
                    &\multicolumn{3}{c|}{Boost}&\multicolumn{3}{c|}{Suppress}&\multicolumn{3}{c|}{Boost}&\multicolumn{3}{c}{Suppress}\cr
				\midrule 
				Model&Related& Unrelated &\textcolor{brandeisblue}{$\Uparrow$} Ratio& Related& Unrelated&\textcolor{brandeisblue}{$\Uparrow$} Ratio&Related& Unrelated &\textcolor{brandeisblue}{$\Uparrow$} Ratio& Related& Unrelated&\textcolor{brandeisblue}{$\Uparrow$} Ratio\cr
				\midrule
                    Random Neurons &-0.03&-0.03&1.0&+0.06&+0.11&0.55&+0.08&+0.04&2.0&-0.01&-0.01&1.0\cr
                    Activation &+92.53&+91.73&1.0&-45.44&-45.14&1.0&+44.17&+40.28&1.1&-31.04&-28.88&1.1\cr
                    Knowledge Neurons$^*$~\shortcite{dai2022knowledge} &+932.05&+921.84&1.0&-85.70&-85.34&1.0&+1081.33&+161.98&6.7&-86.74&-48.18&1.8\cr
                    QRNCA \textit{wo/} ICA&+2403.60&+982.52&2.5&-82.82&-74.09&1.1&+1225.27&+190.03&6.5&-81.62&-36.93&2.2\cr
                    QRNCA \textit{w/} Common Neurons&+919.03&+328.49&2.8&-59.34&-33.59&1.8&+606.54&+54.84&10.4&-71.45&-8.40&8.5\cr
				\rowcolor{teal!5} \textbf{QRNCA} &+77.23&+17.55&\textbf{4.4}&-27.65&-4.95&\textbf{5.6}&+248.64&+6.91&\textbf{41.2}&-56.20&+1.56&\textbf{36.0}\cr
				\bottomrule
			\end{tabular}
			\caption{Details of average probability percentage changes of related and unrelated queries.  The LLM here is Llama-2-7B~\cite{touvron2023llama}} 	\label{tab:detail_prob_impact}
		\end{threeparttable}
	}
	%\end{minipage}%
	\vspace{-10pt}
\end{table*}

\begin{table*}[ht] 
	\centering This research was partially supported by  the UKRI INDICATE project (Grant No. EP/Y017749/1), by the ERC under the EU’s Horizon 2020 research and innovation programme (grant agreement No. 101020934), and by J.P. Morgan and the Royal Academy of Engineering under the Research Chairs and Senior Research Fellowships scheme.
	\small
	\setlength{\tabcolsep}{2.2mm}{
		\begin{threeparttable} 
			\begin{tabular}{l p{12.5cm}}  
				\toprule  
				\textbf{Prompt ID}&\textbf{Template}\cr
				\midrule
				\textit{Domain Prompt 1}&\texttt{You will be asked a multiple-choice question. Respond with the letter which corresponds to the correct answer, followed by a period. There is no need to provide an explanation, so your response should be very short.\textbackslash nNow here is the question:\textbackslash n\texttt{\{Question\}}\textbackslash n A. \texttt{\{A\}}\textbackslash n B. \texttt{\{B\}}\textbackslash n C. \texttt{\{C\}}\textbackslash n D. \texttt{\{D\}}\textbackslash nResponse:}\cr
				\midrule
				\textit{Domain Prompt 2}&\texttt{Prepare to answer a multiple-choice question. Provide the letter that corresponds to the correct answer, followed by a period. Keep your response brief; no explanations are necessary.\textbackslash nHere is the question:\textbackslash n\texttt{\{Question\}}\textbackslash n A. \texttt{\{A\}}\textbackslash n B. \texttt{\{B\}}\textbackslash n C. \texttt{\{C\}}\textbackslash n D. \texttt{\{D\}}\textbackslash nResponse:}\cr
				\midrule
    \textit{Domain Prompt 3}&\texttt{Below is a multiple-choice question. Respond with the letter that best answers the question. Keep your response brief, stating only the letter corresponding to your answer, followed by a period, with no explanation.\textbackslash nThe question is:\textbackslash n\texttt{\{Question\}}\textbackslash n A. \texttt{\{A\}}\textbackslash n B. \texttt{\{B\}}\textbackslash n C. \texttt{\{C\}}\textbackslash n D. \texttt{\{D\}}\textbackslash nResponse:}\cr
    \midrule
    \textit{Language Prompt 1}&\texttt{You will be asked a multiple-choice question. Respond with the letter which corresponds to the correct answer, followed by a period. There is no need to provide an explanation, so your response should be very short. \textbackslash nNow here is the question:\textbackslash n\texttt{\{Question\}} \textbackslash nHere the [Y] is most likely to be?
\textbackslash n A. \texttt{\{A\}}\textbackslash n B. \texttt{\{B\}}\textbackslash n C. \texttt{\{C\}}\textbackslash n D. \texttt{\{D\}}\textbackslash nResponse:}\cr
\midrule
    \textit{Language Prompt 2}&\texttt{Prepare to answer a multiple-choice question. Provide the letter that corresponds to the correct answer, followed by a period. Keep your response brief; no explanations are necessary. \textbackslash nNow here is the question:\textbackslash n\texttt{\{Question\}} \textbackslash nHere the [Y] is most likely to be?
\textbackslash n A. \texttt{\{A\}}\textbackslash n B. \texttt{\{B\}}\textbackslash n C. \texttt{\{C\}}\textbackslash n D. \texttt{\{D\}}\textbackslash nResponse:}\cr
\midrule
\textit{Language Prompt 3}&\texttt{Below is a multiple-choice question. Respond with the letter that best answers the question. Keep your response brief, stating only the letter corresponding to your answer, followed by a period, with no explanation. \textbackslash nNow here is the question:\textbackslash n\texttt{\{Question\}} \textbackslash nHere the [Y] is most likely to be?
\textbackslash n A. \texttt{\{A\}}\textbackslash n B. \texttt{\{B\}}\textbackslash n C. \texttt{\{C\}}\textbackslash n D. \texttt{\{D\}}\textbackslash nResponse:}\cr
				\bottomrule
			\end{tabular}
			\caption{Prompt templates for constructing multi-choice QA datasets. We use ChatGPT to translate English templates to other languages. } 	\label{tab:prompt}
		\end{threeparttable}
	}
	%\end{minipage}%
\end{table*}

\begin{figure*}[t]%
	\centering
	\subfloat[\centering Boosting QR neurons of domains]{{\includegraphics[width=0.5\textwidth]{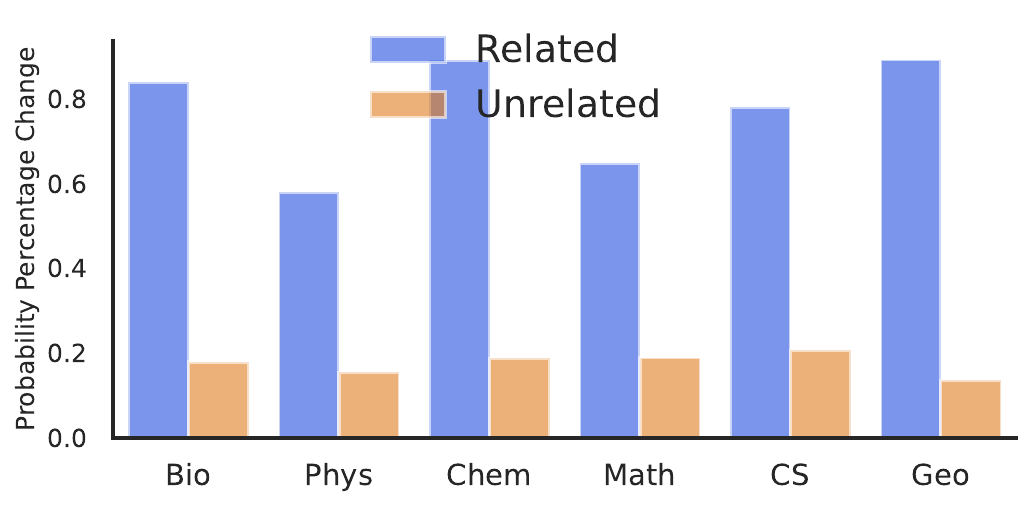} }}%
	%\qquad
	\subfloat[\centering Suppressing QR neurons of domains]{{\includegraphics[width=0.5\textwidth]{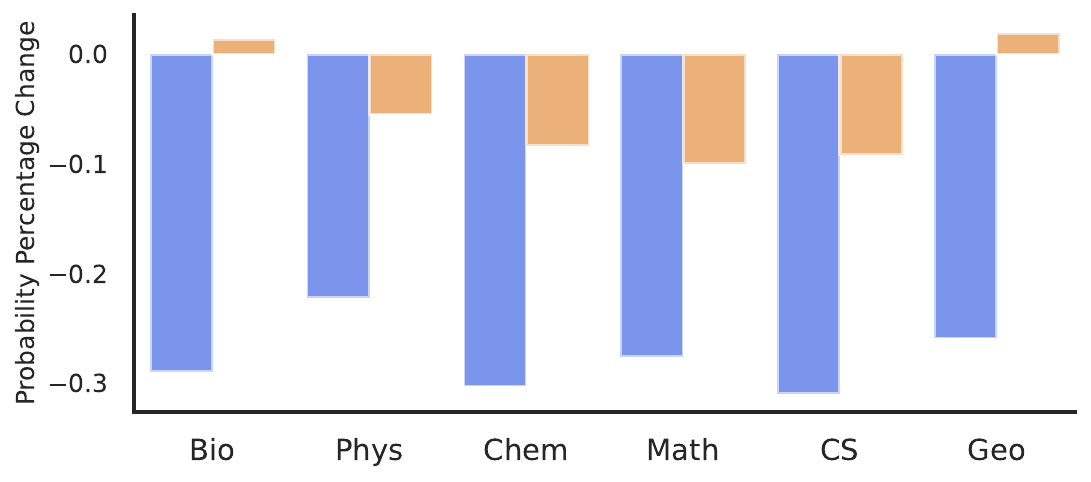} }}%
	\caption{The correct probability percentage change across different domains. The LLM here is \textbf{\textcolor{red}{Mistral-7B}}~\cite{jiang2023mistral} }
	\label{fig:mistral_prob_change_domain}%
 \vspace{-10pt}
\end{figure*}

\begin{figure*}[t]%
	\centering
	\subfloat[\centering Boosting QR neurons of domains]{{\includegraphics[width=0.5\textwidth]{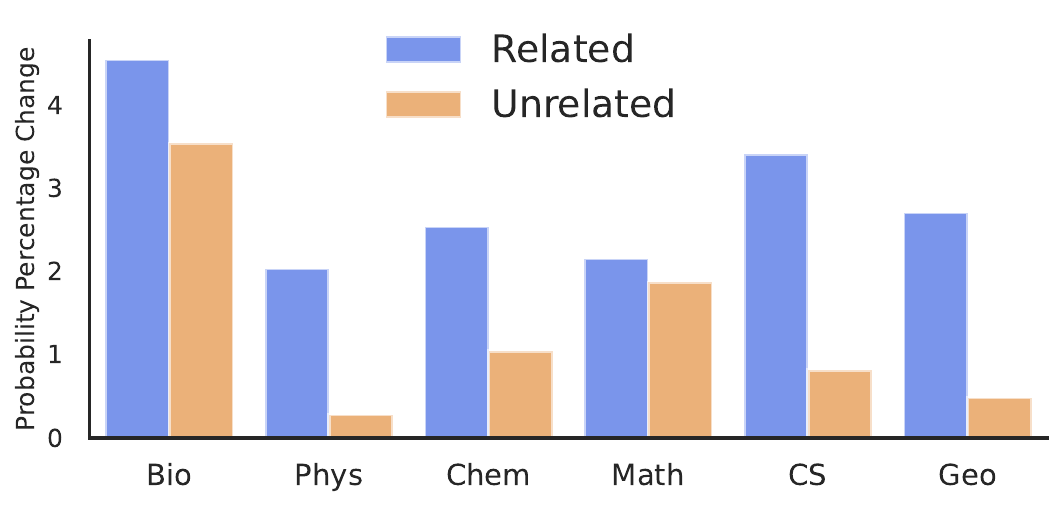} }}%
	%\qquad
	\subfloat[\centering Suppressing QR neurons of domains]{{\includegraphics[width=0.5\textwidth]{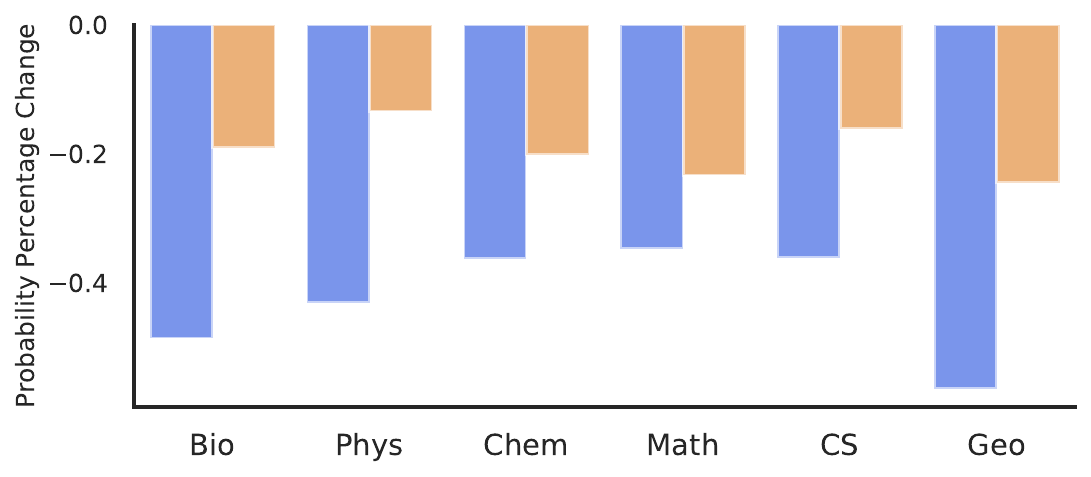} }}%
	\caption{An ablation study of using \textbf{$\frac{\partial P_{q}}{\partial g}$} to compute naica scores. The LLM here is Llama-2-7B~\cite{touvron2023llama}. }
	\label{fig:gate_prob_change_domain}%
 \vspace{-10pt}
\end{figure*}

\begin{table*}[ht] 
	\centering 
	\tiny
	\setlength{\tabcolsep}{2.0mm}{
		\begin{threeparttable} 
			\begin{tabular}{c|p{7cm}|p{7cm}}  
				\toprule  
				\textbf{Field}&\textbf{Question}&\textbf{Options}\cr
				\midrule
				Biology&\textit{The energy given up by electrons as they move through the electron transport chain is used to?}&\texttt{ \makecell{A. make glucose\\ B. make NADH\\ \textcolor{red}{C. produce ATP}\\ D. break down glucose}}\cr
                \midrule
				Physics&\textit{An object is placed 100 cm from a plane mirror. How far is the image from the object?}& \texttt{\makecell{ A. 50 cm\\ \textcolor{red}{B. 200 cm}\\ C. 100 cm\\ D. 300 cm}}\cr
                \midrule
				Chemistry&\textit{Three half-lives after an isotope is prepared:}& \texttt{\makecell{ A. 12.5\% of the isotope decayed\\ B. 25\% of the isotope decayed\\ C. 25\% of the isotope is left\\ \textcolor{red}{D. 12.5\% of the isotope is left}}}\cr
                \midrule
				Mathematics&\textit{Suppose the graph of f is both increasing and concave up on a <= x <= b. Then, using the same number of subdivisions, and with L, R, M, and T denoting, respectively, left, right, midpoint, and trapezoid sums, it follows that:}& \texttt{\makecell{ A. R <= T <= M <= L\\ \textcolor{red}{B. L <= M <= T <= R}\\ C. R <= M <= T <= L\\ D. L <= T <= M <= R}}\cr
                \midrule
				Computer Science&\textit{A programmer is writing a program that is intended to be able to process large amounts of data. Which of the following considerations is LEAST likely to affect the ability of the program to process larger data sets?}& \texttt{\makecell{ A. How long the program takes to run\\ \textcolor{red}{B. How many programming statements the program contains}\\ C. How much storage space the program requires as it runs\\ D. How much memory the program requires as it runs}}\cr
                \midrule
				Geography&\textit{The tendency for migration to decrease with distance is called?}& \texttt{\makecell{ A. push factors.\\ B. migration selectivity.\\ \textcolor{red}{C. distance decay.}\\ D. pull factors.}}\cr
                \midrule
				English&\textit{Sergey Lavrov was born in [Y]. Here the [Y] is most likely to be?}& \texttt{\makecell{A. Montevideo\\ B. Bengaluru\\ C. Parsons\\ \textcolor{red}{D. Moscow}}}\cr
				\bottomrule
			\end{tabular}
			\caption{Examples in our constructed datasets. For the language dataset, we only show one English example as multilingual samples are obtained bu using traslator~\cite{kassner2021multilingual} }\label{tab:data_examples}
		\end{threeparttable}
	}
\end{table*}

\end{document}